\documentclass[runningheads]{llncs}

\usepackage{accv}

\usepackage{accvabbrv}

\usepackage[pagebackref,breaklinks,colorlinks,citecolor=accvblue]{hyperref}
\usepackage{graphicx,verbatim}
\usepackage{booktabs}
\usepackage{amssymb}
\usepackage{pifont}
\usepackage{amsmath}
\usepackage{xcolor}
\usepackage[table]{xcolor}
\usepackage{algorithm}
\usepackage{algpseudocode}
\usepackage{enumitem}
\usepackage{orcidlink}

\DeclareUnicodeCharacter{25CF}{$\bullet$}
\DeclareUnicodeCharacter{25CB}{$\circ$}

\newcommand{\st}{\scriptsize}
\newcommand{\plm}{$\pm$}

\usepackage[accsupp]{axessibility}  % Improves PDF readability for those with disabilities.

\usepackage{orcidlink}

\begin{document}

\title{Mixture of Multi-Sequence Feature Difference for Contrast-Enhanced MRI Synthesis}
\titlerunning{Mixture of Feature Difference for CE-MRI Synthesis}


\author{Jonghun Kim \orcidlink{0009-0002-2790-2090} \and
Junmo Kwon \orcidlink{0000-0002-2242-1637} \and
Hyunjin Park \thanks{Corresponding Author} \orcidlink{0000-0001-5681-8918}}

\authorrunning{J. Kim et al.}

\institute{Department of Electrical and Computer Engineering, \\ Sungkyunkwan University, Suwon, Korea \\
\email{\{iproj2,skenfn1231,hyunjinp\}@skku.edu}}

\maketitle

\begin{abstract}
Contrast-enhanced magnetic resonance imaging (CE-MRI) is crucial for accurately diagnosing and segmenting brain tumors. However, the acquisition of CE-MRI requires injecting contrast agents, which poses potential health risks and increases scan costs. Here, we propose a CE-MRI synthesis method from non-contrast sequences, reducing both risk and cost. Unlike existing methods that rely on channel-wise concatenation to fuse multi-sequence information for synthesis, we introduce a Mixture of Feature Difference (MoFD) module. The proposed module uses a dynamic weighting strategy to adaptively select and aggregate informative features from both individual source sequences and explicit differences among them.
Quantitative and qualitative evaluations demonstrate that our proposed method outperforms baselines in terms of image fidelity and perceptual quality on the UCSF-PDGM and UPenn-GBM datasets. Our method successfully reconstructs tumor regions and achieves segmentation performance comparable to that obtained with real CE-MRI. Our ablation studies show that although adversarial learning improves perceptual quality, a regression-based variant of our method achieves higher structural fidelity.  Our code available at \href{https://github.com/jongdory/MoFD}{github.com/jongdory/MoFD}.


\end{abstract}
\section{Introduction}

\begin{figure}[t]
    \centering
    \includegraphics[width=0.9\textwidth]{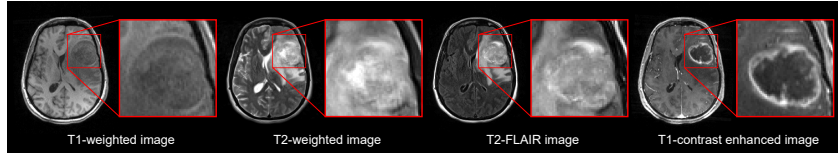}
    \vspace{-6pt}
    \caption{Illustration of non-contrast multi-sequence MRI and contrast-enhanced MRI. Red bounding boxes show zoomed-in views of the tumor region. Unlike the non-contrast sequence images, the T1 contrast-enhanced image clearly delineates tumor boundaries.}
    \label{intro}
    \vspace{-12pt}
\end{figure}

Contrast-enhanced magnetic resonance imaging (CE-MRI) is a pivotal tool that uses a contrast agent injected into the bloodstream to enhance image clarity \cite{weinmann1984characteristics,caravan1999gadolinium, kim2024radiomics, kim2025simulating}. 
Although non-contrast MRI is sufficient for observing anatomical structures, it often lacks the sensitivity needed to identify vascular conditions and subtle tissue discrepancies. It is difficult to distinguish tumors or inflammation from healthy tissue on non-contrast scans because these conditions often have similar intensities \cite{carr1984intravenous}. CE-MRI addresses this issue by using contrast agents that make blood vessels and lesions hyperintense on T1-weighted images. This hyperintensity enables the precise identification of the location and extent of these lesions \cite{lohrke201625}.
As shown in Fig. \ref{intro}, tumor boundaries are ambiguous in other non-contrast MRI sequences, whereas the contrast-enhanced image clearly delineates the tumor region.
Despite these advantages, administering contrast agents raises safety concerns because gadolinium may not be fully excreted, potentially leading to accumulation in the brain and bones \cite{kanda2014high,murata2016macrocyclic}. Furthermore, CE-MRI requires longer acquisition times and incurs higher costs.

To address these issues, recent studies have explored generative model-based medical image translation for synthesizing CE-MRI from non-CE images \cite{kim2026posterior}.
In particular, Generative Adversarial Network (GAN) \cite{goodfellow2014generative} and diffusion model \cite{ho2020denoising} based generative models have been widely investigated for synthesizing target modalities by concatenating source modalities along the channel dimension as input \cite{yu2019ea, sharma2019missing, chen2022synthesizing, dalmaz2022resvit, kim2024adaptive}.
Ea-GAN \cite{yu2019ea} is a 3D Pix2Pix \cite{isola2017image}-based model that integrates edge information, effectively preserving structural details. MM-GAN \cite{sharma2019missing} utilizes a 3D GAN framework that combines information from all available MRI sequences to simultaneously synthesize arbitrary missing sequences. 
Chen et al. \cite{chen2022synthesizing} have proposed a 3D convolutional network designed using a fusion method that extracts unique information feature maps through independent encoders for each input modality and then combines them in the channel direction to synthesize CE-MRI within the regression framework.
ResViT \cite{dalmaz2022resvit} has been introduced as a GAN-based generative model for multi-modal image synthesis that combines Vision Transformers \cite{dosovitskiy2020image} with the convolutional networks to leverage global context. ALDM \cite{kim2024adaptive} applies a 3D Latent Diffusion Model \cite{rombach2022high} incorporating switchable SPADE \cite{park2019semantic} blocks within the latent space, enabling the translation of a single source image into various target images. 
Recent studies \cite{rassmann2026regression} have suggested that, in medical image translation, fidelity is more critical than perceptual quality because generative models for restoration tasks are subject to the perception-distortion tradeoff \cite{blau2018perception}. Accordingly, regression losses can be more important than adversarial learning, which primarily aims to improve perceptual quality \cite{rassmann2026regression}.

Most approaches concatenate multi-sequence images along the channel dimension for use as input. This often overlooks the distinct contrast characteristics of each sequence.
For example, T1-weighted images depict fat as hyperintense, while water, cerebrospinal fluid (CSF), edema, and inflammation appear dark \cite{pooley2005fundamental,ginat2012intracranial}. T2-weighted images depict water, lesions, and fat as hyperintense, while air and blood appear hypointense \cite{pooley2005fundamental,bitar2006mr}. T2-FLAIR images also highlight lesions and fat, but water and CSF appear dark \cite{de1992mr}. Furthermore, the contrast enhancement of pathologies such as inflammation and edema varies significantly across these sequences, serving as an important biomarker \cite{cho2024quantification}. 
These abnormal regions can be explicitly modeled using difference maps between sequences, making them useful for accurate contrast-enhanced image synthesis. 
Therefore, to accurately synthesize contrast-enhanced regions, it is crucial to explicitly incorporate the diverse contrast variations of distinct sequences and their explicit differences rather than relying on channel-wise concatenation.
Because channel-wise concatenation captures pixel-level relationships between source modalities, it can be sensitive to intensity scaling and noise. In contrast, modeling differences in a feature space can provide an enhanced contextual representation of inter-sequence contrast changes.

We propose the Mixture of Feature Difference (MoFD) module that effectively leverages the complementary information from multiple sequences for CE-MRI synthesis. Unlike conventional methods that concatenate sequences across channels, our approach uses an encoder to extract sequence-specific features from each input sequence capturing unique contrast characteristics. We further calculate the feature differences between sequences and dynamically allocate importance to sequence-specific features and their feature differences, enabling flexible and selective feature integration instead of fixed averaging. We validate our method with regression-based and GAN-based approaches to investigate the impact of adversarial training on improving perceptual quality. We also evaluate tumor segmentation performance on the synthesized images using nnU-Net \cite{isensee2021nnu} to assess synthesis quality in tumor regions.

\noindent \textbf{Our Contributions}:
\begin{itemize}[noitemsep, topsep=0pt, partopsep=0pt, parsep=0pt]
\item[•] We propose a method that explicitly leverages feature differences for multi-sequence MRI, enabling the model to capture distinct contrast variations across MRI sequences.
\item[•] We introduce the MoFD module to prioritize informative features via adaptive and selective integration.
\item[•] We analyze the impact of regression-based and GAN-based objectives on CE-MRI synthesis.
\item[•] We extensively validate our method on UCSF-PDGM and UPENN-GBM datasets for CE-MRI synthesis. 
\end{itemize}
\section{Related Work}

\subsection{Medical Image Translation for CE-MRI Synthesis}
Gadolinium-based contrast agents improve lesion visibility but raise safety concerns, motivating recent studies on CE-MRI synthesis from non-contrast sequences.
However, most existing methods still rely on channel-wise concatenation of input modalities, which can overlook the distinct contrast characteristics of each sequence. Bone et al. \cite{bone2021contrast} synthesized full-dose T1 contrast-enhanced (T1ce) brain MRI from either non-contrast images or low-dose contrast-enhanced images to reduce concerns related to gadolinium retention and nephrotoxicity \cite{ledneva2009renal}. Their study considered a clinical scenario in which T1, T2-FLAIR, and apparent diffusion coefficient (ADC) images were acquired before contrast injection, followed by a 25\% low-dose T1ce scan and an additional 75\% injection for the full-dose T1ce reference. The source modalities were concatenated along the channel dimension and used as input.
Chen et al. \cite{chen2022synthesizing} proposed a 3D fully convolutional network to synthesize CE-MRI from T1, T2, and ADC images. Although regression-based methods can preserve anatomical structures, they often produce over-smoothed images and struggle to synthesize fine details in localized enhancing lesions. To improve perceptual quality, Jiao et al. \cite{jiao2023contrast} proposed a GAN-based model for synthesizing abdominal contrast-enhanced T1ce images from non-contrast T1 and T2 images, using separate encoders and a sparse attention fusion module to combine modality-specific features. 
Li et al. \cite{li2024evaluating} demonstrated that virtual contrast-enhanced MRI can precisely delineate tumors for nasopharyngeal carcinoma radiotherapy, eliminating the risk of gadolinium-based contrast agent.
Bottani et al. \cite{bottani2024contrast} compared three conditional GAN-based generators for translating non-contrast MRI into CE-MRI, each built on a 3D U-Net backbone with residual connections, attention gates, or transformer layers.
Most existing methods rely on GAN-based frameworks. However, due to the perception-distortion trade-off \cite{blau2018perception}, adversarial learning tends to only improve perceptual image quality at the cost of reduced fidelity. This trade-off is important in medical image translation, where fidelity is more critical than perceptual quality. Accordingly, regression losses can be more important than adversarial learning, which primarily aims to improve perceptual quality \cite{rassmann2026regression}. 

\subsection{Feature Fusion and Difference Modeling}
Medical image translation has been studied to reduce scan time, cost, and patient risk by generating missing or target modalities from acquired sequences \cite{dalmaz2022resvit,kim2025tumor,kim2026visual}. Early methods used convolutional networks and GANs to translate non-contrast sequences into target images, while later studies introduced more advanced fusion strategies to combine complementary information across modalities. For example, Hi-Net \cite{zhou2020hi} used hybrid feature fusion to capture modality-specific and cross-modal information. Han et al. \cite{han2023explainable, han2024synthesis} introduced a task-specific fusion method that adaptively weights each modality according to its contribution to the target image.
Rokuss et al. \cite{rokuss2024longitudinal} introduced a difference weighting block to explicitly incorporate visual and temporal changes between baseline and follow-up scans into the model architecture. Their method computed feature-map differences between the two time points and uses them to generate an attention map that emphasizes changed regions.
Mixture-of-Experts (MoE) \cite{shazeer2017} is an effective module for scaling model capacity while keeping the number of activated parameters during inference relatively small, often leading to improved performance. Oh et al. \cite{oh2025distributionaware} interpreted MoE from a control-theoretic perspective, formulating it as a feedback control mechanism that dynamically assigns the influence of each controller according to the current state to optimize the overall system.
Building on these prior studies, we propose a module for CE-MRI synthesis that explicitly computes feature-map differences across modalities and uses a gating network to dynamically select informative features from multi-sequence MRI.
\section{Method}
\begin{figure}[t]
    \centering
    \includegraphics[width=0.99\textwidth]{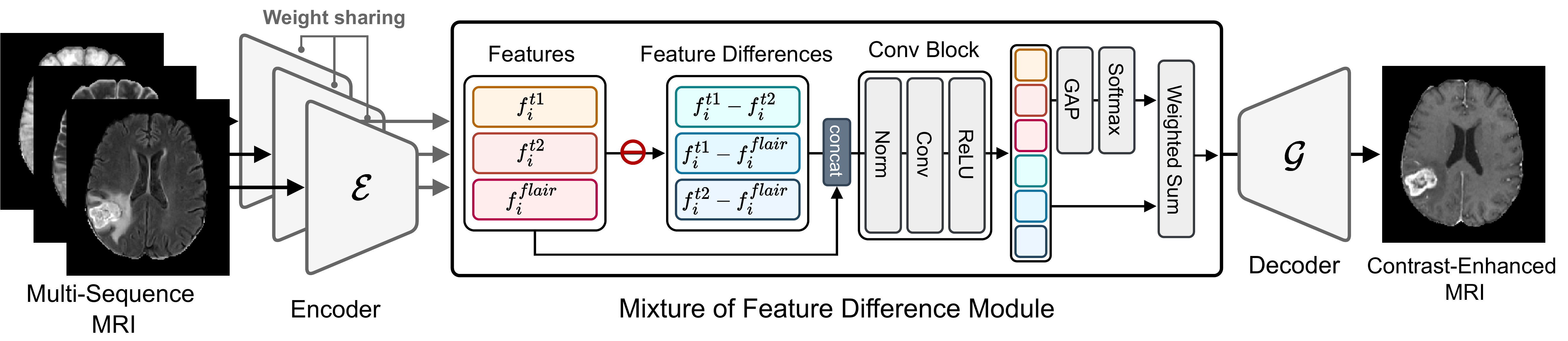}
    % \vspace{-3pt}
    \caption{Illustration of the proposed method for CE-MRI synthesis. Weight-sharing encoders extract features from each input sequence (T1, T2, and T2-FLAIR). The model uses the differences between these features and employs a softmax function to assign adaptive weights dynamically. These weighted features are then aggregated and fed into the decoder to synthesize the CE-MRI.}
    \label{fig1}
\end{figure}

Our model architecture is based on the U-Net \cite{ronneberger2015u}, as illustrated in Fig. \ref{fig1}. Unlike previous approaches that concatenate multi-sequence images along the channel dimension and process them through a single encoder, our model processes each sequence individually to extract distinct features. This strategy allows for the explicit separation of sequence-specific features, which are then utilized for contrast-enhanced image synthesis.
We employ a weight-sharing encoder across all input sequences. This design ensures that different input sequences are mapped into a unified latent feature space under a consistent criterion, while maintaining parameter efficiency \cite{liu2017unsupervised,liu2016coupled}. 

\subsection{Mixture of Feature Difference}
We compute both the encoder features and feature differences and pass their mixture to the decoder.
Let $S = \{I_1, I_2, \dots, I_N\}$ be the set of input sequences, where $N$ is the number of sequences and $I^k$ denotes the image of the $k$-th sequence. The encoder extracts multi-scale features from each $I^k$. To capture the inter-sequence contrast variations, we define the feature difference set $D_i$ at the $i$-th scale as the collection of pairwise differences between sequence features:
\begin{equation}
\text{Encoder}(I^{k}) = \{f^{k}_1, f^{k}_2, \dots, f^{k}_L\}, \quad D_i = \{ f^k_i - f^l_i \mid 1 \le k < l \le N \},
\label{eq:1}
\end{equation}
where $f^k_i$ represents the latent feature of the $k$-th sequence at the $i$-th scale and $L$ is the total number of feature scales.

Instead of simply averaging the features defined in Eq. \ref{eq:1} and passing them to the decoder, we design a mechanism to dynamically select and prioritize significant features. To this end, we compute importance scores for each feature map and forward a weighted sum of these features to the decoder.
Both the individual features of a given sequence and the feature differences between sequences are first processed through a convolution block with pre-normalization to ensure consistent scaling. 
We concatenate the feature set $F_i$ and the feature difference set $D_i$ to form the combined representation $C_i$. This is then processed through a convolution block to yield the refined feature map $\tilde{F}_i$, as formulated in:
\begin{equation}
C_i = \text{Concat}(F_i, D_i), \ \ \ \ \tilde{F}_i = \text{Conv}(C_i).
\label{eq:2}
\end{equation}

We compute the attention weights $a_i$ by applying Global Average Pooling (GAP) followed by a softmax.
The final aggregated feature $\bar{f}_i$, which is forwarded to the decoder, is obtained as a weighted sum of these features:
\begin{equation}
A_i = \text{Softmax}(\text{GAP}(\tilde{F}_i)), \ \ \ \ \bar{f}_i = \sum_{m=1}^M a^m_i \cdot \tilde{f}^m_i,
\label{eq:3}
\end{equation}
where $a^m_i \in A_i$ denotes the weight assigned to the $m$-th feature at the $i$-th scale and $M$ is the total number of elements in the combined set, defined as the sum of the number of elements in $F_i$ and $D_i$.
The aggregated multi-scale features are then passed to the decoder to synthesize the CE-MRI. Let $\bar{F}=\{\bar{f}_1, \bar{f}_2, \dots, \bar{f}_L\}$ denote the set of aggregated features across all scales. The prediction $\hat{Y}$ is defined as follows: 
\begin{equation}
\hat{Y} = \text{Decoder}(\bar{F}) = \text{Decoder}(\bar{f}_1, \bar{f}_2, \dots, \bar{f}_L).
\end{equation}
Rather than merging all input sequences at the image level, the weight-sharing encoder first maps each sequence into a common feature space, enabling the model to extract comparable representations, while preserving the distinctive contrast information of each modality. The MoFD module then exploits both individual encoder features and their pairwise differences to identify complementary information for contrast-enhanced image synthesis. By adaptively weighting these features at multiple scales, the decoder receives a refined representation that emphasizes relevant cross-modality cues while reducing the influence of redundant or less informative features. This design is particularly effective for multi-sequence MRI synthesis, where the target contrast is determined not by a single input modality but by complex relationships among anatomical structure, tissue contrast, and lesion appearance across sequences. 

\subsection{Structural Alignment}
\begin{figure}[t]
    \centering
    \includegraphics[width=\textwidth]{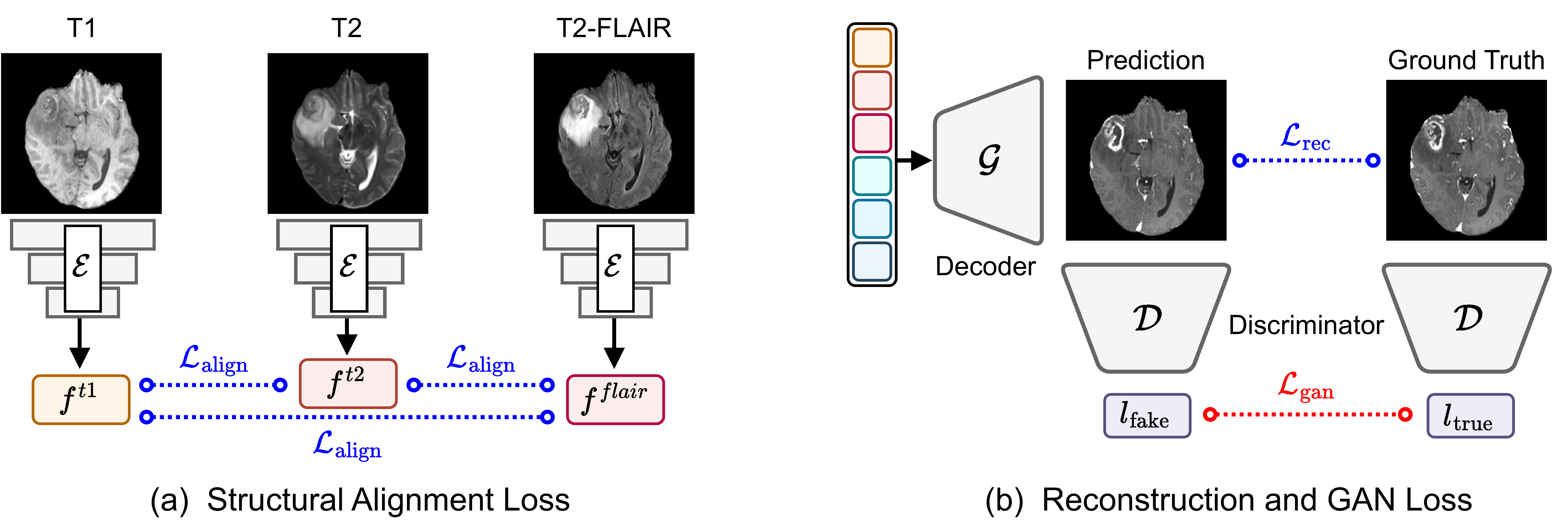}
    \caption{Illustration of the training objectives. (a) The structural alignment loss encourages structural information in sequence features to remain consistent across input sequences. (b) The reconstruction loss encourages the prediction to match the ground truth, while the optional GAN loss promotes realistic textures in the synthesized image.}
    \label{trainloss}
\end{figure}

Although different MRI sequences exhibit varying contrast, they share the underlying anatomical structure. Therefore, we aim to enforce structural consistency across multiple sequences for a single subject. Inspired by previous work \cite{xiong2025learning}, we focus on the structure comparison component of the Structural Similarity Index Measure (SSIM) \cite{wang2004image} for the sequence feature $f^k_i$. We exclude the luminance and contrast terms from the original SSIM to focus on preserving structural alignment despite intensity variations between sequences. We compute localized statistics using a Gaussian kernel. 
The alignment loss is applied to all pairwise combinations of sequence features, as shown in Fig. \ref{trainloss} (a). 
The alignment loss for structural similarity is defined as follows:
\begin{equation}
s(f^k_i, f^l_i) = \frac{\sigma{(f^k_i, f^l_i)}+ \epsilon}{\sigma{(f^k_i)} \sigma{(f^l_i)} + \epsilon}, \ \ \ \ \mathcal{L}_\text{align}= 1 - s(f^k_i, f^l_i),
\label{eq:4}
\end{equation}
where $\sigma(f^k_i, f^l_i)$ denotes the covariance between $f^k_i$ and $f^l_i$, $\sigma(f^k_i)$ and $\sigma(f^l_i)$ represent their respective standard deviations, and $\epsilon = 10^{-6}$ is a small constant. The structural alignment loss encourages features from different sequences of the same subject to maintain consistent anatomical structure, which helps the model preserve patient-specific structure during synthesis. 

\subsection{Training Strategy}
To ensure high-fidelity synthesis of CE-MRI, we optimize the model using a reconstruction loss. This objective encourages the predicted image $\hat{Y}$ to closely match the ground truth $Y$:
\begin{equation}
\mathcal{L}_\text{rec} = ||Y - \hat{Y}||_1.
\label{eq:5}
\end{equation}
Additionally, we incorporate the structural consistency constraint defined in Eq. \ref{eq:4} to enforce structural coherence across different sequences. 
The total objective function is formulated as:
\begin{equation}
\mathcal{L}_\text{total} = \mathcal{L}_\text{rec} + \mathcal{L}_\text{align}.
\end{equation}
Although using a GAN loss can improve the perceptual quality of synthesized images, it is subject to the perception-distortion trade-off \cite{blau2018perception}. In medical imaging, prioritizing perceptual quality can result in hallucinated details or artifacts that do not exist in the ground truth, compromising pixel-level fidelity. Therefore, as shown in Fig. \ref{trainloss} (b), we define the adversarial loss using a PatchGAN \cite{isola2017image} discriminator $\mathcal{D}$ as follows, treating it as an optional component:
\begin{equation}
\mathcal{L}_\text{gan} = \mathbb{E}_{Y}[\log \mathcal{D}(Y)] + \mathbb{E}_{\hat{Y}}[\log(1 - \mathcal{D}(G(S))],
\label{eq:6}
\end{equation}
where $S$ is set of input sequences. The total objective function with the GAN term is formulated as:
\begin{equation}
\mathcal{L}_\text{total} = \mathcal{L}_\text{rec} + \mathcal{L}_\text{align} + \mathcal{L}_\text{gan}.
\label{eq:7}
\end{equation}
These components allow the proposed method to balance structural fidelity and contrast generation.
The reconstruction objective promotes voxel-level accuracy, the alignment objective regularizes anatomical consistency, and the optional adversarial objective can be used when improved perceptual realism is desired. 
\section{Experiments}

\subsection{Datasets.}
We used two public datasets: UCSF-PDGM \cite{calabrese2022university,calabrese2022university2} and UPenn-GBM \cite{bakas2021multi,bakas2022university}. For the experiments, the UCSF-PDGM dataset was divided into 210 subjects for training, 23 for validation, and 54 for testing. The UPenn-GBM dataset was divided into 210 training, 21 validation, and 52 testing subjects. Both datasets consist of multi-sequence MRI, including T1, T2, T2-FLAIR, and T1ce images. In addition, both datasets provide glioblastoma tumor masks annotated into whole tumor (WT), enhancing tumor (ET), and tumor core (TC) regions. All volumes were preprocessed with co-registration to the same anatomical template, skull stripping, and resampling to an isotropic resolution of $1\times1\times1\text{ mm}^3$.

\subsection{Implementation details.}
The model was trained for 500 epochs using the Adam \cite{kingma2014adam} optimizer with a batch size of 1 and a learning rate of 1e-4. For training and inference, input images were cropped to minimize background regions and adjusted to a spatial resolution of $192\times192\times128$. Our model was built on a 3D U-Net \cite{ronneberger2015u} backbone, with feature channel dimensions set to [32, 64, 128, 256, 512] at each stage. The proposed MoFD module was inserted into the skip connections at all stages, allowing encoder features to be refined before being passed to the corresponding decoder layers.

\subsection{Metrics.}
To evaluate image quality, the 3D PSNR and SSIM metrics were used. Perceptual quality was assessed using the LPIPS \cite{zhang2018unreasonable} metric with a 3D ResNet-50 \cite{he2016deep} trained on RadImageNet \cite{mei2022radimagenet}. For downstream evaluation, we trained an nnU-Net \cite{isensee2021nnu} on single-input T1ce images from the training set and evaluated tumor segmentation performance using the Dice score for the WT, ET, and TC regions. 

\subsection{Comparison methods.}
We compared our method with representative MRI synthesis models spanning both 3D and 2D architectures. Among the 3D methods, Ea-GAN \cite{yu2019ea} is based on a 3D pix2pix \cite{isola2017image}, Chen et al. \cite{chen2022synthesizing} is a regression-based model and ALDM \cite{kim2024adaptive} employs a 3D latent diffusion model \cite{rombach2022high}. In contrast, MM-GAN \cite{sharma2019missing} and ResViT \cite{dalmaz2022resvit} are 2D GAN-based methods that operate on axial images. HiNet \cite{zhou2020hi} and MMgSN-Net \cite{li2024evaluating} are also 2D GAN-based approaches and they incorporate modality fusion modules to integrate information from multiple input modalities.

\begin{figure}[t]
    \centering
    \includegraphics[width=0.95\textwidth]{figs_rebuttal/result_UCSF.pdf}
    \vspace{-6pt}
    \caption{\textbf{Qualitative results of T1ce synthesis with comparison methods} on the UCSF-PDGM \cite{calabrese2022university,calabrese2022university2}. The bounding box (bottom left) shows a zoomed-in view of the tumor region from the T1ce. 
    The bottom right image shows the corresponding tumor segmentation results by nnU-Net \cite{isensee2021nnu}.
    Without GAN, the results are blurry. With GAN, however, the perceptual quality improves. }
    \label{result_UCSF}
    \vspace{-3pt}
    \setlength{\tabcolsep}{2pt} 
    \centering
    \captionof{table}{\textbf{Quantitative performance of T1ce synthesis with comparison methods} on the UCSF-PDGM \cite{calabrese2022university,calabrese2022university2}. PSNR and SSIM are measured and reported as 3D metrics in mean $\pm$ std. Bold indicates the \textbf{best}, underlined denotes the \underline{second-best} performance. WT: Whole Tumor, ET: Enhancing Tumor, TC: Tumor Core.}
    \vspace{3pt}
    \label{table1}
    \centering
    \scalebox{0.675}{
        \begin{tabular}{l|ccc|cccc}
            \toprule
            \textbf{Dataset} & \multicolumn{7}{c}{\textbf{UCSF-PDGM} \cite{calabrese2022university,calabrese2022university2}}   \\
            \cmidrule(lr){1-1} \cmidrule(lr){2-4} \cmidrule(lr){5-8}
            \textbf{Tasks} & \multicolumn{3}{c|}{\textbf{T1ce Synthesis}} & \multicolumn{4}{c}{\textbf{Tumor Segmentation}} \\
            \cmidrule(lr){1-1} \cmidrule(lr){2-4} \cmidrule(lr){5-8} 
            \textbf{Model} & \textbf{PSNR} $\uparrow$ & \textbf{SSIM} $\uparrow$  & \textbf{LPIPS} $\downarrow$ & \textbf{WT} & \textbf{ET} & \textbf{TC} & \textbf{Avg.} \\
            \cmidrule(lr){1-8} 
            Ea-GAN \cite{yu2019ea}                  &            23.59 \st\plm\ 0.99 &            0.872 \st\plm\ 0.029  & 0.047 \st\plm\ 0.022 & 0.844 \st\plm\ 0.069 & 0.318 \st\plm\ 0.257 & 0.125 \st\plm\ 0.217 & 0.429 \st\plm\ 0.161 \\
            MM-GAN \cite{sharma2019missing}         &            24.24 \st\plm\ 1.43 &            0.877 \st\plm\ 0.033  & 0.042 \st\plm\ 0.025 & 0.747 \st\plm\ 0.140 & 0.094 \st\plm\ 0.186 & 0.046 \st\plm\ 0.137 & 0.296 \st\plm\ 0.117 \\
            Chen et al. \cite{chen2022synthesizing} &            24.54 \st\plm\ 1.38 &            0.880 \st\plm\ 0.027  & 0.025 \st\plm\ 0.016 & \underline{0.852 \st\plm\ 0.093} & 0.711 \st\plm\ 0.195 & 0.532 \st\plm\ 0.268 & 0.698 \st\plm\ 0.160 \\
            ResViT \cite{dalmaz2022resvit}          &            24.46 \st\plm\ 1.34 &            0.871 \st\plm\ 0.026  & 0.025 \st\plm\ 0.020 & \underline{0.852 \st\plm\ 0.079} & 0.741 \st\plm\ 0.140 & 0.551 \st\plm\ 0.200 & 0.714 \st\plm\ 0.120 \\
            ALDM   \cite{kim2024adaptive}           &            23.65 \st\plm\ 0.88 &            0.848 \st\plm\ 0.026  & 0.025 \st\plm\ 0.017 & 0.803 \st\plm\ 0.108 & 0.679 \st\plm\ 0.190 & 0.443 \st\plm\ 0.252 & 0.642 \st\plm\ 0.141 \\
            Hi-Net   \cite{zhou2020hi}              &            24.12 \st\plm\ 1.45 &            0.874 \st\plm\ 0.036  & 0.029 \st\plm\ 0.017 & 0.826 \st\plm\ 0.102 & 0.581 \st\plm\ 0.257 & 0.410 \st\plm\ 0.277 & 0.606 \st\plm\ 0.193 \\
            MMgSN-Net \cite{li2024evaluating}       &            22.59 \st\plm\ 1.94 &            0.858 \st\plm\ 0.095  & 0.049 \st\plm\ 0.027 & 0.842 \st\plm\ 0.090 & 0.662 \st\plm\ 0.270 & 0.519 \st\plm\ 0.266 & 0.674 \st\plm\ 0.187 \\
            \rowcolor{gray!10} \textbf{MoFD} \st(w/o GAN)              &    \textbf{25.00 \st\plm\ 1.26} &    \textbf{0.894 \st\plm\ 0.027} & \underline{0.024 \st\plm\ 0.020} & \textbf{0.867 \st\plm\ 0.080} & \textbf{0.769 \st\plm\ 0.165} & \textbf{0.588 \st\plm\ 0.214} & \textbf{0.741 \st\plm\ 0.134} \\
            \rowcolor{gray!10} \textbf{MoFD} \st(w GAN)                & \underline{24.62 \st\plm\ 1.19} & \underline{0.881 \st\plm\ 0.029} &    \textbf{0.022 \st\plm\ 0.017} & 0.821 \st\plm\ 0.104 & \underline{0.763 \st\plm\ 0.146} & \underline{0.559 \st\plm\ 0.174} & \underline{0.714 \st\plm\ 0.107} \\
            \bottomrule
        \end{tabular}
    }
    \vspace{-9pt}
\end{figure}
\section{Results}

\begin{figure}[t]
    \centering
    \includegraphics[width=0.95\textwidth]{figs_rebuttal/result_UPENN.pdf}
    \vspace{-6pt}
    \caption{\textbf{Qualitative results of T1ce synthesis with comparison methods} on the UPENN-GBM \cite{bakas2021multi,bakas2022university}. The bounding box (bottom left) shows a zoomed-in view of the tumor region from the T1ce. 
    The bottom right image shows the corresponding tumor segmentation results by nnU-Net \cite{isensee2021nnu}.}
    \label{result_UPENN}
    \vspace{-3pt}
    \setlength{\tabcolsep}{2pt} 
    \centering
    \captionof{table}{\textbf{Quantitative performance of T1ce synthesis with comparison methods} on the UPENN-GBM \cite{bakas2021multi,bakas2022university}. PSNR and SSIM are measured and reported as 3D metrics in mean $\pm$ std. Bold indicates the \textbf{best}, underlined denotes the \underline{second-best} performance.}
    \vspace{3pt}
    \label{table2}
    \centering
    \scalebox{0.675}{
        \begin{tabular}{l|ccc|cccc}
            \toprule
            \textbf{Dataset} & \multicolumn{7}{c}{\textbf{UPENN-GBM} \cite{bakas2021multi,bakas2022university}}   \\
            \cmidrule(lr){1-1} \cmidrule(lr){2-4} \cmidrule(lr){5-8}
            \textbf{Tasks} & \multicolumn{3}{c|}{\textbf{T1ce Synthesis}} & \multicolumn{4}{c}{\textbf{Tumor Segmentation}} \\
            \cmidrule(lr){1-1} \cmidrule(lr){2-4} \cmidrule(lr){5-8} 
            \textbf{Model} & \textbf{PSNR} $\uparrow$ & \textbf{SSIM} $\uparrow$  & \textbf{LPIPS} $\downarrow$ & \textbf{WT} & \textbf{ET} & \textbf{TC} & \textbf{Avg.} \\
            \cmidrule(lr){1-8} 
            Ea-GAN \cite{yu2019ea}                  &            23.08 \st\plm\ 0.82  &            0.833 \st\plm\ 0.026  & 0.047 \st\plm\ 0.022 & 0.726 \st\plm\ 0.141 & 0.454 \st\plm\ 0.276 & 0.213 \st\plm\ 0.218 & 0.464 \st\plm\ 0.181 \\
            MM-GAN \cite{sharma2019missing}         &            23.82 \st\plm\ 1.22  &            0.840 \st\plm\ 0.029  & 0.036 \st\plm\ 0.022 & 0.689 \st\plm\ 0.186 & 0.570 \st\plm\ 0.356 & 0.312 \st\plm\ 0.259 & 0.524 \st\plm\ 0.242 \\
            Chen et al. \cite{chen2022synthesizing} & \underline{24.35 \st\plm\ 1.25} & \underline{0.842 \st\plm\ 0.026} & 0.025 \st\plm\ 0.016 & 0.616 \st\plm\ 0.218 & 0.618 \st\plm\ 0.333 & 0.372 \st\plm\ 0.316 & 0.535 \st\plm\ 0.247 \\
            ResViT \cite{dalmaz2022resvit}          &            24.28 \st\plm\ 1.15  &            0.828 \st\plm\ 0.027  & 0.025 \st\plm\ 0.014 & 0.718 \st\plm\ 0.237 & \underline{0.652 \st\plm\ 0.327} & \underline{0.473 \st\plm\ 0.289} & 0.614 \st\plm\ 0.256 \\
            ALDM   \cite{kim2024adaptive}           &            23.55 \st\plm\ 0.77  &            0.814 \st\plm\ 0.024  & 0.026 \st\plm\ 0.017 & 0.665 \st\plm\ 0.232 & 0.646 \st\plm\ 0.325 & 0.412 \st\plm\ 0.282 & 0.574 \st\plm\ 0.252 \\
            Hi-Net   \cite{zhou2020hi}              &            23.49 \st\plm\ 1.01  &            0.845 \st\plm\ 0.026  & 0.026 \st\plm\ 0.018 & 0.718 \st\plm\ 0.189 & 0.550 \st\plm\ 0.307 & 0.353 \st\plm\ 0.299 & 0.541 \st\plm\ 0.243 \\
            MMgSN-Net \cite{li2024evaluating}       &            22.16 \st\plm\ 0.91  &            0.810 \st\plm\ 0.029  & 0.029 \st\plm\ 0.018 & 0.628 \st\plm\ 0.312 & 0.543 \st\plm\ 0.372 & 0.344 \st\plm\ 0.309 & 0.505 \st\plm\ 0.303 \\
            \rowcolor{gray!10} \textbf{MoFD} \st(w/o GAN)              &    \textbf{24.80 \st\plm\ 1.10} &    \textbf{0.861 \st\plm\ 0.025} & \underline{0.024 \st\plm\ 0.015} & \textbf{0.763 \st\plm\ 0.137} & \textbf{0.721 \st\plm\ 0.277} & \textbf{0.520 \st\plm\ 0.262} & \textbf{0.668 \st\plm\ 0.206} \\
            \rowcolor{gray!10} \textbf{MoFD} \st(w GAN)                &            24.33 \st\plm\ 1.01  &            0.839 \st\plm\ 0.026  & \textbf{0.022 \st\plm\ 0.014} & \underline{0.760 \st\plm\ 0.148} & 0.626 \st\plm\ 0.371 & 0.464 \st\plm\ 0.332 & \underline{0.616 \st\plm\ 0.264} \\ 
            \bottomrule
        \end{tabular}
    }
    \vspace{-9pt}
\end{figure}

\subsection{Results on the Comparison Methods}
We evaluated the performance of T1ce synthesis against baseline models quantitatively (Table \ref{table1}) and qualitatively (Fig. \ref{result_UCSF}) on the UCSF-PDGM \cite{calabrese2022university,calabrese2022university2} dataset. 
Most baseline models were trained using T1, T2, and T2-FLAIR sequences fused via channel-wise concatenation, whereas Hi-Net and MMgSN-Net used the multi-modal fusion strategies proposed in their respective studies.
As shown in Table \ref{table1}, our regression-based method (MoFD w/o GAN) achieved the best fidelity performance, with the highest PSNR and SSIM scores. The second-best fidelity results were obtained by Chen et al. \cite{chen2022synthesizing}, a regression-based model for T1ce synthesis. In general, models trained with adversarial loss are subject to the perception-distortion trade-off \cite{blau2018perception}: they can improve perceptual quality, as reflected by lower LPIPS scores, but often reduce pixel-level fidelity in terms of PSNR and SSIM. Despite this tendency, our GAN-based method (MoFD w/ GAN) remained competitive, achieving the best LPIPS score and the second-best PSNR and SSIM scores on UCSF-PDGM.
Qualitative comparisons in Fig. \ref{result_UCSF} further support these quantitative findings. While several comparison methods fail to capture detailed tumor appearance, our method reconstructs both anatomical structures and contrast-enhanced tumor regions more faithfully. In particular, Hi-Net struggles to delineate the tumor core region, whereas MMgSN-Net fails to adequately represent the edema component within the whole tumor region. These qualitative observations are consistent with the segmentation results in Table \ref{table1}, where the proposed model achieves the best Dice scores for WT, ET, TC, and their average, indicating that its synthesized T1ce images preserve tumor structures more accurately than the comparison methods. A similar trend was observed on the UPENN-GBM dataset, as shown in Fig. \ref{result_UPENN} and Table \ref{table2}.

\subsection{Multi-Sequence Combinations}
We investigated the impact of different input sequence combinations on synthesis quality and downstream segmentation performance. We used an nnU-Net \cite{isensee2021nnu} trained on ground truth T1ce images to segment the synthesized T1ce images. Table \ref{table3} presents the quantitative comparison.
The model using all sequences (T1, T2, and T2-FLAIR) achieved the highest scores for fidelity, perceptual quality, and downstream segmentation. For segmentation, the model showed high performance for both WT and ET. However, the model showed limited performance on the TC, the smallest of the subregions.
Note that although using only T1 as an input generally improves image fidelity due to its structural consistency with T1ce, it leads to poor segmentation results. This discrepancy arises because T1 images provide insufficient contrast information for distinguishing tumor features. As illustrated in Fig. \ref{result_input}, the qualitative results confirm that the multi-sequence input enables the model to reconstruct tumor details.

\begin{figure}[t]
    \centering
    \includegraphics[width=0.99\textwidth]{figs_rebuttal/result_input.pdf}
    \vspace{-6pt}
    \caption{\textbf{Qualitative results of T1ce synthesis with various input sequence combinations} on the UCSF-PDGM \cite{calabrese2022university,calabrese2022university2}. The bounding boxes display a zoomed-in view of the boxed area in the bottom left. The bottom right image shows the corresponding tumor segmentation results by nnU-Net \cite{isensee2021nnu}.}
    \label{result_input}
    \setlength{\tabcolsep}{3pt} 
    \centering
    \captionof{table}{\textbf{Quantitative results of T1ce synthesis and segmentation with various input sequence combinations} on the UCSF-PDGM \cite{calabrese2022university,calabrese2022university2}. }
    \vspace{3pt}
    \label{table3}
    \centering
    \scalebox{0.675}{
        \begin{tabular}{ccc|ccc|cccc}
            \toprule
             \multicolumn{3}{c}{\textbf{Modalities}} & \multicolumn{3}{|c}{\textbf{T1ce Synthesis}} & \multicolumn{4}{|c}{\textbf{Tumor Segmentation}}  \\
            \cmidrule(lr){1-3} \cmidrule(lr){4-6} \cmidrule(lr){7-10} 
            $\text{T}_1$ & $\text{T}_2$ & $\text{T}_{2f}$ & \textbf{PSNR} $\uparrow$ & \textbf{SSIM} $\uparrow$ & \textbf{LPIPS} $\downarrow$ & \textbf{WT} & \textbf{ET} & \textbf{TC} & \textbf{Avg.} \\
            \cmidrule(lr){1-10} 
            ● & ○ & ○ & 23.770 \st\plm\ 1.485 & 0.857 \st\plm\ 0.045 &  0.031 \st\plm\ 0.020 & 0.788 \st\plm\ 0.115 & 0.674 \st\plm\ 0.176  & 0.410 \st\plm\ 0.287 & 0.624 \st\plm\ 0.153 \\
            ○ & ● & ○ & 23.723 \st\plm\ 0.942 & 0.868 \st\plm\ 0.025 &  0.032 \st\plm\ 0.016 & 0.808 \st\plm\ 0.091 & 0.724 \st\plm\ 0.151  & 0.385 \st\plm\ 0.224 & 0.639 \st\plm\ 0.116 \\
            ○ & ○ & ● & 22.660 \st\plm\ 0.890 & 0.843 \st\plm\ 0.027 &  0.037 \st\plm\ 0.019 & 0.815 \st\plm\ 0.088 & 0.728 \st\plm\ 0.134  & 0.221 \st\plm\ 0.224 & 0.588 \st\plm\ 0.118 \\
            ● & ● & ○ & 24.360 \st\plm\ 1.225 & 0.880 \st\plm\ 0.028 &  0.024 \st\plm\ 0.014 & 0.807 \st\plm\ 0.096 & 0.723 \st\plm\ 0.143  & 0.496 \st\plm\ 0.178 & 0.675 \st\plm\ 0.110 \\
            ● & ○ & ● & 23.678 \st\plm\ 1.283 & 0.859 \st\plm\ 0.034 &  0.025 \st\plm\ 0.015 & 0.813 \st\plm\ 0.089 & 0.738 \st\plm\ 0.134 & 0.480 \st\plm\ 0.184 & 0.677 \st\plm\ 0.100 \\
            ○ & ● & ● & 24.091 \st\plm\ 0.864 & 0.877 \st\plm\ 0.022 &  0.025 \st\plm\ 0.013 & 0.801 \st\plm\ 0.093 & 0.724 \st\plm\ 0.131  & 0.445 \st\plm\ 0.212 & 0.657 \st\plm\ 0.104 \\
            \rowcolor{gray!10} ● & ● & ● & \textbf{24.626 \st\plm\ 1.197} & \textbf{0.881 \st\plm\ 0.029} & \textbf{0.022 \st\plm\ 0.017} & \textbf{0.821 \st\plm\ 0.104} & \textbf{0.763 \st\plm\ 0.146} & \textbf{0.559 \st\plm\ 0.174} & \textbf{0.714 \st\plm\ 0.107} \\
            \cmidrule(lr){1-10}
            \multicolumn{3}{c|}{GT T1ce} & inf & 1.00 & 0 & 0.826 \st\plm\ 0.095 & 0.767 \st\plm\ 0.121 & 0.840 \st\plm\ 0.121 & 0.811 \st\plm\ 0.088  \\
            \bottomrule
        \end{tabular}
    }
\end{figure}

\subsection{Ablation Study}

\noindent \textbf{MoFD Architecture.} Table \ref{table4} details the ablation results regarding the architecture of the proposed module (Fig. \ref{fig1}). Omitting the convolution block leads to inconsistent feature scales due to the lack of normalization, which negatively affects the synthesis quality. Using a weighted sum to calculate the final output significantly outperforms the approach of averaging features, confirming the effectiveness of our dynamic feature selection mechanism.

\vspace{3pt}
\noindent \textbf{Loss Function.} Table \ref{table5} presents the ablation study on the three components of $\mathcal{L}_\text{total}$ (Eq. \ref{eq:7}). Excluding the $\mathcal{L}_\text{gan}$ yielded superior fidelity scores in T1ce synthesis. While incorporating the $\mathcal{L}_\text{gan}$ improved perceptual quality, it reduced fidelity. Notably, incorporating $\mathcal{L}_\text{align}$ improved both fidelity and quality metrics in all cases. Interestingly, the regression-based model (w/o $\mathcal{L}_\text{gan}$) demonstrated better tumor segmentation performance than the GAN-based model. This indicates that although adversarial training improves perceptual realism, it risks introducing realistic but fake details that deviate from the ground truth, thereby degrading segmentation accuracy. Thus, for tasks requiring structural precision, a regression-based approach might be better.

\vspace{3pt}
\noindent \textbf{Top-K Selection.} Table \ref{table6} summarizes the performance of the Top-K feature selection strategy, which is similar to the MoE \cite{shazeer2017} framework. We evaluated the model by varying $K$ to select the top-weighted features. The results indicate that the best performance was achieved when all six available features, comprising three sequence features and three difference maps, across both datasets was better than selecting a partial subset.

\begin{table} [t]
    \setlength{\tabcolsep}{2pt} 
    \centering
    \caption{\textbf{Ablation study on the MoFD Architecture.} }
    \label{table4}
    \centering
    \scalebox{0.67}{
        \begin{tabular}{ccc|ccc|cccc}
            \toprule
            \multicolumn{3}{c|}{\textbf{Dataset}} & \multicolumn{7}{c}{\textbf{UCSF-PDGM} \cite{calabrese2022university,calabrese2022university2}}   \\
            \cmidrule(lr){1-3} \cmidrule(lr){4-6} \cmidrule(lr){7-10} 
            \multicolumn{3}{c|}{\textbf{Tasks}} & \multicolumn{3}{c|}{\textbf{T1ce Synthesis}} & \multicolumn{4}{c}{\textbf{Tumor Segmentation}}  \\
            \cmidrule(lr){1-3} \cmidrule(lr){4-6} \cmidrule(lr){7-10} 
            \multicolumn{3}{c|}{\textbf{Method}} & \textbf{PSNR} $\uparrow$ & \textbf{SSIM} $\uparrow$ & \textbf{LPIPS} $\downarrow$ & \textbf{WT} & \textbf{ET} & \textbf{TC} & \textbf{Avg.} \\
            \cmidrule(lr){1-10} 
            \multicolumn{3}{c|}{w/o Conv Block}  &         23.590 \st\plm\ 1.059  &         0.859 \st\plm\ 0.026  & 0.028 \st\plm\ 0.014 & 0.820 \st\plm\ 0.097 & 0.709 \st\plm\ 0.131 & 0.501 \st\plm\ 0.185 & 0.677 \st\plm\ 0.093 \\
            \multicolumn{3}{c|}{Average Feature} &         24.301 \st\plm\ 1.198  &         0.875 \st\plm\ 0.029  & 0.023 \st\plm\ 0.019 & \textbf{0.824 \st\plm\ 0.094} & 0.716 \st\plm\ 0.146 & 0.512 \st\plm\ 0.175 & 0.685 \st\plm\ 0.098 \\
            \cmidrule(lr){1-10} 
            \rowcolor{gray!10} \multicolumn{3}{c|}{\textbf{Ours}}   & \textbf{24.626 \st\plm\ 1.197} & \textbf{0.881 \st\plm\ 0.029} & \textbf{0.022 \st\plm\ 0.017} & 0.821 \st\plm\ 0.104 & \textbf{0.763 \st\plm\ 0.146} & \textbf{0.559 \st\plm\ 0.174} & \textbf{0.714 \st\plm\ 0.107} \\
            \bottomrule
        \end{tabular}
    }
\end{table}

\begin{table} [t]    
    \setlength{\tabcolsep}{2pt} 
    \centering
    \caption{\textbf{Ablation study on the Loss function.}}
    \vspace{-3pt}
    \label{table5}
    \centering
    \scalebox{0.67}{
        \begin{tabular}{ccc|ccc|cccc}
            \toprule
            \multicolumn{3}{c|}{\textbf{Dataset}} & \multicolumn{7}{c}{\textbf{UCSF-PDGM} \cite{calabrese2022university,calabrese2022university2}}   \\
            \cmidrule(lr){1-3} \cmidrule(lr){4-6} \cmidrule(lr){7-10} 
            \multicolumn{3}{c|}{\textbf{Tasks}} & \multicolumn{3}{c|}{\textbf{T1ce Synthesis}} & \multicolumn{4}{c}{\textbf{Tumor Segmentation}} \\
            \cmidrule(lr){1-3} \cmidrule(lr){4-6} \cmidrule(lr){7-10} 
            $\mathcal{L}_{rec}$ & $\mathcal{L}_{align}$ & $\mathcal{L}_{gan}$ & \textbf{PSNR} $\uparrow$ & \textbf{SSIM} $\uparrow$ & \textbf{LPIPS} $\downarrow$ & \textbf{WT} & \textbf{ET} & \textbf{TC}  & \textbf{Avg.} \\
            \cmidrule(lr){1-10} 
            \checkmark                    &            &            &         24.833 \st\plm\ 1.263  &         0.891 \st\plm\ 0.029  & 0.027 \st\plm\ 0.014 & 0.859 \st\plm\ 0.078 & 0.757 \st\plm\ 0.117 & 0.474 \st\plm\ 0.264 & 0.696 \st\plm\ 0.115 \\
            \rowcolor{gray!10} \checkmark                    & \checkmark &            & \textbf{25.009 \st\plm\ 1.263} & \textbf{0.894 \st\plm\ 0.027} & 0.024 \st\plm\ 0.020  & \textbf{0.867 \st\plm\ 0.080} & \textbf{0.769 \st\plm\ 0.165} & \textbf{0.588 \st\plm\ 0.214} & \textbf{0.741 \st\plm\ 0.134} \\
             \checkmark &            & \checkmark &         24.025 \st\plm\ 1.211  &         0.888 \st\plm\ 0.026  & 0.023 \st\plm\ 0.013 & 0.824 \st\plm\ 0.092 & 0.718 \st\plm\ 0.145 & 0.545 \st\plm\ 0.152 & 0.695 \st\plm\ 0.098 \\
            \rowcolor{gray!10} \checkmark & \checkmark & \checkmark &         24.626 \st\plm\ 1.197  &         0.881 \st\plm\ 0.029  & \textbf{0.022 \st\plm\ 0.017} & 0.821 \st\plm\ 0.104 & 0.763 \st\plm\ 0.146 & 0.559 \st\plm\ 0.174 & 0.714 \st\plm\ 0.107 \\
            \bottomrule
        \end{tabular}
    }

    \vspace{12pt}
    \setlength{\tabcolsep}{5pt} 
    \centering
    \caption{\textbf{Ablation study on the Top-K Selection}. }
    \label{table6}
    \centering
    \scalebox{0.725}{
        \begin{tabular}{c|ccc|ccc}
            \toprule
            \textbf{Dataset} & \multicolumn{3}{|c}{\textbf{UCSF-PDGM} \cite{calabrese2022university,calabrese2022university2}} & \multicolumn{3}{|c}{\textbf{UPenn-GBM} \cite{bakas2021multi,bakas2022university}}  \\
            \cmidrule(lr){1-1} \cmidrule(lr){2-4} \cmidrule(lr){5-7} 
            \textbf{Top-K} & \textbf{PSNR} $\uparrow$ & \textbf{SSIM} $\uparrow$ & \textbf{LPIPS} $\downarrow$ & \textbf{PSNR} $\uparrow$ & \textbf{SSIM} $\uparrow$ & \textbf{LPIPS} $\downarrow$  \\
            \cmidrule(lr){1-7} 
            1 & 24.270 \st\plm\ 1.199 & 0.875 \st\plm\ 0.031 & 0.025 \st\plm\ 0.015 & 23.902 \st\plm\ 1.044 & 0.830 \st\plm\ 0.027 & 0.023 \st\plm\ 0.019 \\
            2 & 24.294 \st\plm\ 1.202 & 0.875 \st\plm\ 0.031 & 0.024 \st\plm\ 0.015 & 23.940 \st\plm\ 0.961 & 0.831 \st\plm\ 0.027 & 0.023 \st\plm\ 0.016 \\
            3 & 24.476 \st\plm\ 1.181 & 0.880 \st\plm\ 0.030 & 0.023 \st\plm\ 0.013 & 24.001 \st\plm\ 0.955 & 0.833 \st\plm\ 0.027 & 0.022 \st\plm\ 0.017 \\
            4 & 24.461 \st\plm\ 1.192 & 0.880 \st\plm\ 0.030 & 0.023 \st\plm\ 0.014 & 24.092 \st\plm\ 0.973 & 0.835 \st\plm\ 0.026 & 0.022 \st\plm\ 0.016 \\
            5 & 24.437 \st\plm\ 1.202 & 0.878 \st\plm\ 0.030 & 0.024 \st\plm\ 0.015 & 24.202 \st\plm\ 0.982 & 0.836 \st\plm\ 0.027 & 0.022 \st\plm\ 0.017 \\
            \rowcolor{gray!10} 6 & \textbf{24.626 \st\plm\ 1.197} & \textbf{0.881 \st\plm\ 0.029} & \textbf{0.022 \st\plm\ 0.017} & \textbf{24.336 \st\plm\ 1.011} & \textbf{0.839 \st\plm\ 0.026} & \textbf{0.022 \st\plm\ 0.014} \\
            \bottomrule
        \end{tabular}
    }
\end{table}

\begin{figure}[t]
    \centering
    \includegraphics[width=0.95\textwidth]{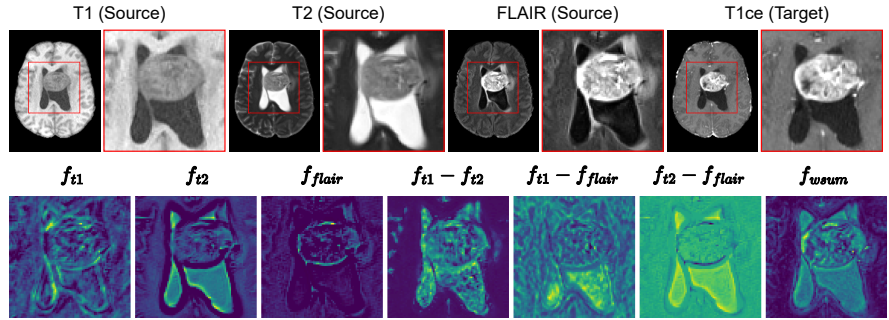}
    \caption{\textbf{Visualization of features, feature differences, and weighted sum features} derived from source T1, T2, and FLAIR modalities. $f_{wsum}$ denotes the weighted sum feature. The red bounding boxes indicate zoomed-in views of the tumor region. }
    \label{feat_vis}
\end{figure}

\begin{figure}[t]
    \includegraphics[width=\textwidth]{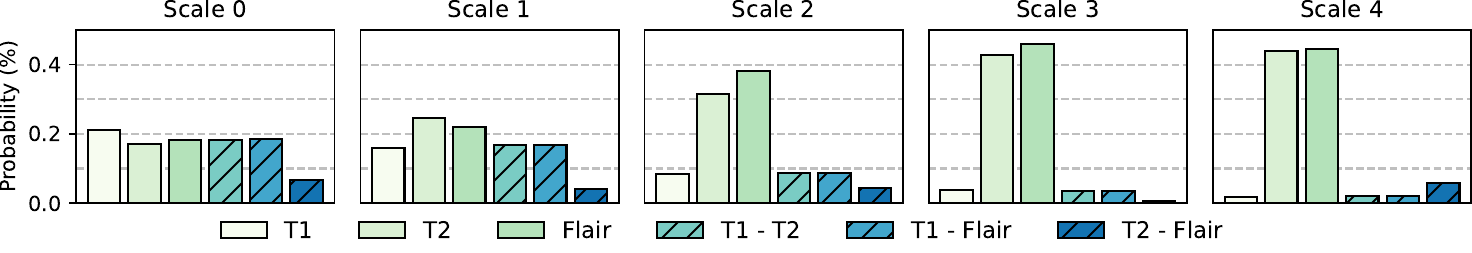}
    \caption{\textbf{Visualization of the feature selection probability in the MoFD module}. At low-level scales (scales 0 and 1), the selection probabilities are relatively evenly distributed across features and feature differences. As the network becomes deeper and approaches higher-level representations, the selection probabilities for T2 and T2-FLAIR features increase. Since image translation requires accurate pixel-level synthesis, low-level features passed through the skip connections remain important and the proposed module effectively uses all available features across scales. }
    \label{feature_select}
\end{figure}

\subsection{Analysis of Feature Selection}

We visualized the features and feature differences produced by the MoFD module, as shown in Fig. \ref{feat_vis}. In the tumor region, each feature and feature difference captures the tumor appearance from a different representation. The weighted-sum feature is then generated by adaptively combining these features according to their selection probabilities. Notably, this aggregated feature closely resembles the tumor region in the target image, suggesting that the proposed module effectively extracts complementary representations and dynamically integrates them to produce a final feature representation suitable for target T1ce synthesis. 

We further visualized which features were selected and passed to the decoder at each scale, as shown in Fig. \ref{feature_select}. At the low-level scales, i.e., scales 0 and 1, the selection probabilities are relatively evenly distributed across the individual features and feature differences, indicating that the MoFD module flexibly selects complementary information to support target synthesis. As the features become deeper, the selection probabilities for T2 and T2-FLAIR tend to increase. This may be because these modalities exhibit contrast patterns similar to T1ce in non-tumor structural regions. However, voxel-level accuracy is particularly important for tumor regions, detailed texture synthesis, and the delineation of enhancing tumor and tumor core regions. Therefore, low-level features can play a more critical role. The relatively balanced feature usage at low-level scales suggests that the proposed module effectively exploits diverse modality-specific and difference-based cues for fine-grained synthesis. This observation is also consistent with Kim et al. \cite{kim2024diffusion}, who showed that coarse layers mainly contribute to forming global image structure, whereas their influence can be reduced when the input image already provides strong structural guidance for image-to-image translation.

\section{Conclusion}
We propose a framework for synthesizing T1ce images from non-contrast multi-sequence MRI using the MoFD module. Instead of relying on simple channel-wise concatenation, MoFD explicitly models both sequence-specific features and their pairwise feature differences, allowing the network to adaptively select complementary information useful for contrast-enhanced image synthesis. Experiments on the UCSF-PDGM and UPenn-GBM datasets showed that our method consistently outperformed baseline approaches in image fidelity and improved downstream tumor segmentation performance. The ablation studies further demonstrated the contribution of the proposed design: feature normalization and dynamic weighting are important for stable fusion, all input sequences and their difference features are beneficial for synthesis quality, and regression-based training can better preserve structural fidelity than adversarial training when accurate tumor representation is required.
% \clearpage  % TODO REVIEW/FINAL: This \clearpage needs to be removed from both review and camera-ready versions.

\section*{Acknowledgement}
This study was supported by the SMC-SKKU Future Convergence Research Program Grant (SMO125123). It was also supported by the National Research Foundation of Korea (NRF) grant (RS-2025-00518533), the AI Graduate School Support Program (Sungkyunkwan University) (RS-2019-II190421), the National Research Foundation of Korea (RS-2024-00408040, RS-2026-25598771), the ICT Creative Consilience Program Grant (IITP-2026-RS-2020-II201821), the Institute of Information \& Communications Technology Planning \& Evaluation (IITP) grant (RS-2026-25528384, Resource-Intensive AI Technologies Based on Sustainable GPU Integrated Platforms), the IITP grant (IITP-2026-RS-2026-25615334, AI Star Fellowship Support Program), and an IITP grant for the AI Computing Support Project for R\&D (RS-2026-25505492, High-Performance Research AI Computing Infrastructure Support at the 2 PFLOPS Scale), all funded by the Ministry of Science and ICT (MSIT), Republic of Korea.

\bibliographystyle{splncs04}
\bibliography{main}
\end{document}